\documentclass[pmlr,nosubfloats]{jmlr}% new name PMLR (Proceedings of Machine Learning)

\usepackage{longtable}% for long tables

 \usepackage{booktabs}
 
\usepackage[load-configurations=version-1]{siunitx} % newer version

\makeatletter
\def\set@curr@file#1{\def\@curr@file{#1}} %temp workaround for 2019 latex release
\makeatother

\theorembodyfont{\upshape}
\theoremheaderfont{\scshape}
\theorempostheader{:}
\theoremsep{\newline}

\usepackage{amsmath,amsfonts,bm}

\def\eqref#1{equation~\ref{#1}}
\def\1{\bm{1}}

\def\vm{{\bm{m}}}

\def\vp{{\bm{p}}}

\def\vs{{\bm{s}}}

\def\vx{{\bm{x}}}
\def\vy{{\bm{y}}}

\def\mW{{\bm{W}}}

\DeclareMathAlphabet{\mathsfit}{\encodingdefault}{\sfdefault}{m}{sl}
\SetMathAlphabet{\mathsfit}{bold}{\encodingdefault}{\sfdefault}{bx}{n}

\def\gL{{\mathcal{L}}}

\newcommand{\R}{\mathbb{R}}

\DeclareMathOperator{\ind}{I}

\def\tpm{{ \(\pm\) }}
\def\vpsi{{\bm{\psi}}}
\def\vomega{{\bm{\omega}}}

\usepackage{hyperref}
\usepackage{url}
\usepackage{multirow}
\usepackage{enumitem}
\usepackage{wrapfig}
\usepackage{subcaption}
\usepackage{algorithm}
\usepackage{algorithmic}

\usepackage{adjustbox}
\usepackage{tikz}
\usepackage{pgfplots}
\usepackage{verbatim}

\usetikzlibrary{arrows.meta}
\usetikzlibrary{positioning}
\usetikzlibrary{calc}
\usetikzlibrary{patterns}
\usetikzlibrary{pgfplots.groupplots}
\pgfplotsset{compat=1.18}

\usetikzlibrary{3d}
\makeatletter % from https://tex.stackexchange.com/a/48776/121799
\tikzoption{canvas is xy plane at z}[]{%
  \def\tikz@plane@origin{\pgfpointxyz{0}{0}{#1}}%
  \def\tikz@plane@x{\pgfpointxyz{1}{0}{#1}}%
  \def\tikz@plane@y{\pgfpointxyz{0}{1}{#1}}%
  \tikz@canvas@is@plane
}
\makeatother

\newcommand{\ModelAbb}{\text{DuETT}}
\newcommand{\modelabb}{\text{DuETT}}
\newcommand{\Modelabb}{\text{DuETT}}

\newcommand{\layername}{\text{DuETT}}

\jmlrvolume{219}
\jmlryear{2023}
\jmlrworkshop{Machine Learning for Healthcare}

\title[DuETT: Dual Event Time Transformer for Electronic Health Records]{DuETT: Dual Event Time Transformer for Electronic Health Records}

\author{\Name{Alex Labach}
\addr Layer 6 AI
       \Email{alex@layer6.ai}
       \AND
       \Name{Aslesha Pokhrel}
       \addr Layer 6 AI, University of Toronto
       \AND
       \Name{Xiao Shi Huang}
\addr Layer 6 AI
\AND
\Name{Saba Zuberi}
\addr Layer 6 AI
\AND
\Name{Seung Eun Yi}
\addr Meta (work done while at Layer 6 AI)
\AND
\Name{Maksims Volkovs}
\addr Layer 6 AI
\AND
\Name{Tomi Poutanen}
\addr Signal 1
\AND
\Name{Rahul G. Krishnan}
\addr University of Toronto \& The Vector Institute
       }

\begin{document}

\maketitle

\begin{abstract}
Electronic health records (EHRs) recorded in hospital settings typically contain a wide range of numeric time series data that is characterized by high sparsity and irregular observations. Effective modelling for such data must exploit its time series nature, the semantic relationship between different types of observations, and information in the sparsity structure of the data. Self-supervised Transformers have shown outstanding performance in a variety of structured tasks in NLP and computer vision. But multivariate time series data contains structured relationships over two dimensions: time and recorded event type, and straightforward applications of Transformers to time series data do not leverage this distinct structure. The quadratic scaling of self-attention layers can also significantly limit the input sequence length without appropriate input engineering. We introduce the \layername{} architecture, an extension of Transformers designed to attend over both time and event type dimensions, yielding robust representations from EHR data. \layername{} uses an aggregated input where sparse time series are transformed into a regular sequence with fixed length; this lowers the computational complexity relative to previous EHR Transformer models and, more importantly, enables the use of larger and deeper neural networks. When trained with self-supervised prediction tasks, that provide rich and informative signals for model pre-training, our model outperforms state-of-the-art deep learning models on multiple downstream tasks from the MIMIC-IV and PhysioNet-2012 EHR datasets.
\end{abstract}

\section{Introduction}

Electronic health record (EHR) data collected in hospitals contains vital sign measurements, lab results, diagnoses, treatments and outcomes.
This multivariate numeric time series is high-dimensional, sparse, and irregularly distributed across time, making it challenging to apply standard time series analysis methods designed for densely sampled data. 
Robust models of clinical outcomes need to leverage the structural characteristics of EHR data. The irregularity and sparsity of observations over time contain valuable information about treatment choices and the evolution of the patient's state. The number of recorded events contain semantic information about the working clinical hypotheses that a clinician has formed about a patient. In this work we present an architecture that explicitly captures this structure of EHR data in both time and event dimensions. 

Our work is motivated by the startling success of Transformer architectures in modelling structured data across a variety of domains. Transformer models currently produce state-of-the-art results on natural language processing (NLP)~\citep{brown2020language}, computer vision~\citep{he2022masked}, and cross-modal learning~\citep{radford2021learning}. But on tabular EHR data~\citep{li2021hi,ren2021rapt,tipirneni2022self}, there remain questions on how to best exploit the structure in such data.
Since computer vision and NLP Transformer models are typically applied over a single dimension of interest, such as position in an image or order of words in text, naively applying these approaches across the time dimension of EHR data means losing information along the event dimension. This can limit the model's ability to capture important relationships between different event types. To mitigate this issue, we propose an extension of the Transformer layer, a \textbf{Du}al \textbf{E}vent \textbf{T}ime \textbf{T}ransformer (DuETT), designed to attend over both time and event dimensions to produce robust representations for EHR data. By adapting the model architecture and training scheme to capture relationships across both dimensions, we can achieve considerably higher accuracy on multiple downstream tasks.

There is a tradeoff between how much finely sampled data is relevant to learn good representations for a predictive task at hand. Prior work has embedded input sequences using a single sequence element for every patient event~\citep{li2021hi,tipirneni2022self}. Although precise, this approach is tricky to scale since memory and runtime complexity of self-attention layers scales quadratically with input length and patients can have hundreds of events in a relatively short period of time. Efficient attention mechanisms~\citep{wang2020linformer, bolya2022hydra} have been proposed, typically trading efficiency for some performance drop, but state-of-the-art Transformer models still generally use quadratic attention due to implementation simplicity and better capacity utilization. Training large models with this sequential EHR input representation consequently requires significant hardware resources or aggressive input truncation, which can negatively impact accuracy. We leverage time binning, which aggregates information and limits the model's computational complexity based on user-selected input granularity.

Applications of deep learning to EHR data face the challenge of having much smaller labelled datasets than typically available in other domains, which can cause severe overfitting in large models. Self-supervised learning (SSL)~\citep{chopra2005contrastive, Caron2021Apr} has risen in popularity as a tool to reduce the dependence of deep learning on large amounts of labelled data, especially for Transformer models. Models are typically pre-trained with SSL using \emph{pseudo-tasks} that are selected to produce robust representations without the need for explicit labels. Pre-trained models are then fine-tuned in a supervised fashion for downstream tasks. The premise of SSL is attractive for EHR data, where few positive samples can be observed for a desired outcome, and privacy limitations can prevent the collection of larger labelled datasets~\citep{Krishnan2022Aug, Bak2022}. We develop SSL training schemes that focuses on learning useful clinical priors of patient state by leveraging the dual event/time representation of EHR data. Doing so provides robust regularization, and enables the training of larger models which when fine-tuned, leads to better accuracy.

In summary, our contributions are (1) the novel \modelabb{} architecture design, which extends Transformers to exploit both time and event modalities of EHR data. (2) The design of an input representation for this architecture that incorporates event information including frequency and missingness, uses early fusion of static variables (age, sex, etc.), and aggregates observations in a way that enables deeper Transformer-based model to be used. (3) A novel self-supervised training scheme that performs masked modelling of measured event values and missingness across both time and event dimensions. (4) A thorough empirical evaluation of our approach on the MIMIC-IV~\citep{mimic_iv_2022} and PhysioNet-2012~\citep{silva_2012} hospital EHR datasets, demonstrating state-of-the-art performances (against both neural network-based and XGBoost baselines) on multiple downstream tasks and effective representation learning during pre-training.

\subsection*{Generalizable Insights about Machine Learning in the Context of Healthcare}

% \textit{This section is required, must keep the above title, and should be the final part of your introduction.  In about one paragraph, or 2-4 bullet points, explain what we should learn from reading this paper that might be relevant to other machine learning in health endeavors.}

Our work provides insights applicable to the evaluation of hospital EHR models. As more hospitals hire and build data science teams, we envision the need for models that decouple representation learning from individual prediction tasks. Our work presents a novel Transformer-based self-supervised architecture, which effectively models the complex relationships between medical observation types, achieving state-of-the-art performance on EHR data. We show that DuETT can be effectively trained with limited labelled data, and can be used to generate patient representations without supervised training, both of which brings practical advantages for developing and deploying predictive models within hospitals. Out of the models we evaluate, we find that our model is the only one that outperforms XGBoost on EHR data.

\section{Related Work}

A variety of neural network models have been proposed for supervised learning on sparse irregular multivariate time series data, such as numeric EHR data. Most are based on recurrent neural networks~\citep{hochreiter1997long, cho2014properties} that expect inputs without missingness, so modifications are required to account for sparse data. Various simple binning and imputation schemes have been explored for converting sparse irregular data into dense regular sequences \cite{shukla2020survey}. mTAN~\citep{shukla2021multitime} uses a more advanced attention-based interpolation approach to produce a regular input for an RNN model. Architectural modifications can also be added to allow RNNs to adapt their hidden state appropriately when inputs are missing, as in CT-GRU~\citep{mozer2017discrete} and GRU-D~\citep{che2018recurrent}. Another line of research uses differential equations to model underlying continuous processes that are related to irregularly sampled inputs~\citep{rubanova2019latent,lechner2020learning,kidger2020neural}, but these approaches require the use of differential equation solvers during training and inference, usually making them slower than ordinary neural networks~\citep{shukla2021multitime}. More recently, Raindrop~\citep{zhang2022graphguided} has applied graph neural networks to aggregate observation embeddings, achieving state-of-the-art results on selected datasets/tasks.

The success of Transformer models in NLP makes them an attractive candidate for other tasks involving sequential data. Many Transformer-based models have been proposed for regular time series data~\citep{wen2022transformers}, but there are fewer models that extend them to sparse irregular time series. RAPT~\citep{ren2021rapt} introduces a modified time-aware attention mechanism to deal with irregular inputs. STraTS~\citep{tipirneni2022self} instead embeds every individual observation as triplets of time, variable and value using MLPs, then passes this sequence of observations to a Transformer. This approach suffers from the limitation that Transformer memory usage is quadratic in sequence length, requiring either the sequence to be aggressively truncated or shallow models to be used (the proposed architecture only uses two Transformer layers and embeddings of length 50). A different approach, Hi-BEHRT~\citep{li2021hi}, uses a hierarchical Transformer architecture to be able to process longer input sequences of individual observations.

Transformer models have also been applied to longitudinal EHR data,\citep{rasmy2021med,li2020behrt,zhang2020inprem}, by using the sets of diagnostic codes applied at hospital or doctor visits as their inputs. However, these models are not applicable to the kinds of data we investigate here, since their inputs do not contain numeric measurements and their temporal resolution is limited to one observation per visit.

Transformers attending across multiple dimensions have been applied in other domains, but with important architectural differences from our model. In computer vision, MLP-Mixer~\cite{tolstikhin2021mlp} showed that alternately processing image spatial and channel dimensions was an effective technique. DaViT~\cite{ding2022davit} uses Transformers instead in an alternating manner, performing local attention operations across spatial and channel dimensions. However, these computer vision models create an internal channel dimension without a semantic relation to an input dimension, whereas we design our input embedding and Transformer layers to preserve the relationship of the event dimension to input events. In dense time-series forecasting, TSMixer~\cite{chen2023tsmixer} has shown the promise of mixing across dimensions with MLP models. Recently, Crossformer~\cite{zhang2023crossformer} has introduced a modified attention function to mix across channels in a dense time-series forecasting Transformer model, but does not apply full quadratic attention or feedforward processing across the channel dimension as our model does.

SSL has become an important framework to enable learning useful representations from data without relying on labels. SSL with Transformers has driven recent advances in NLP and computer vision (e.g.~\cite{devlin2019bert,dosovitskiy2021an}), and has clear potential to advance other fields. Different approaches for SSL with numeric EHR data are explored in~\cite{mcdermott2021comprehensive}, but these are applied to a relatively basic GRU model and do not claim to achieve state-of-the-art results. mTAN uses a similar approach to SSL based on reconstructing inputs, incorporating a variational loss into their training procedure, but without a distinct pre-training stage. SSL is used with Transformers in Hi-BEHRT, which applies BYOL~\citep{grill:hal-02869787} to augmentations of EHR time series data. STraTs uses masked value prediction for SSL with Transformers, while RAPT additionally uses a reasonability check to identify corrupted sequences and a contrastive patient similarity task. Our SSL approach instead adds a presence/absence prediction task, which captures meaningful priors of clinicians regarding a patient's state and introduces a time-wise and event-wise masking strategy.      

\section{Methods}

In this section, we first introduce some useful notation; we then describe the input data and how it is processed into a suitable form for \modelabb{}. Next we outline the \modelabb{} architecture and then finish the section by discussing our training approach, which is made up of a self-supervised pre-training stage followed by a fine-tuning stage. We provide an implementation of \modelabb{} at \url{https://github.com/layer6ai-labs/DuETT}.

 \paragraph{Notation}
For a 3-dimensional tensor $\bm{A} \in \R^{d_1 \times d_2 \times d_3}$, we define a 2D slice as $\bm{A}_{i,\cdot,\cdot}$ and a vector within the tensor as $\bm{A}_{i,j, \cdot}$, where the dot represents all elements of the given dimension of the tensor. We define $a_{i,j} := \bm{A}_{i,j,\cdot}\in \R^{d_3}$ as the vector along the third dimension. The tensor $\bm{A}$ can be reshaped by unfolding along a given dimension. To simplify notation we use $\bm{A}_{i,:,:} \in \R^{d_2 \cdot d_3}$ to denote the vector obtained by flattening $\bm{A}$ along the dimensions with colons.

\subsection{Data}

\paragraph{Input Data Structure}

We consider a dataset structure that corresponds to typical EHR records for patient hospital stays. Each patient stay contains a time series of events corresponding to irregular patient observations, such as vitals and lab results, and a set of static variables that do not change over the course of the stay, such as age and sex. 

This can be represented as a sparse irregular time series dataset of the form $\mathcal{D} = \{ (\vs^p, \mW^p, y^p)\}_{p=1}^N$, where each patient stay $p$ is associated with a set of outcomes $y^p$, a vector of static inputs $\vs^p\in \R^{n_\text{static}}$ and a sequence of events $\mW^p =(w^p_1,w^p_2,\cdots,w^p_{n_p})$ of variable length $n_p$. Each event $w^p_i$ is a triplet
containing the event-type, time since start of stay, and value (if applicable); for example, \texttt{[heart\_rate, 5.32 days, 41bpm]}. The number of unique event types across all patient stays is denoted by $n_{e}$. In subsequent sections we omit patient index $p$ for notational simplicity.

\paragraph{Input Binning} We split the full sequence of patient events, $\mW$, into $n_t$ time bins of equal duration. This transforms the irregularly sampled time series of events into regularly sampled data with missing values \citep{Shukla2020ASO}. For each patient stay, we define a binned input matrix $\vx \in \R^{n_e \times n_t}$ where the element $x_{i,j}$ contains a single value representing an aggregation of all observed values of event-type $i$ in time bin $j$.
Possible choices for the aggregation function include the mean value of events, the maximum or minimum, or the last value observed in the time bin. By adapting the number of bins $n_t$, this input representation allows us to effectively control the trade-off between computational complexity and granularity of event information.

Missing values are very meaningful in EHR data as they often indicate a medical decision to not measure a given event type. In our model, elements of \(\vx\) with no observations in the corresponding time bin are set to $0$; additionally, information on the number of observations across time bins is preserved in a tensor $\vm \in \R^{n_e \times n_t}$, where $m_{i,j}$ is the number of observed events of type \(i\) in time bin \(j\). Passing the number of observed events to the model provides useful information on the types of analyses and treatments that clinicians have selected for the patient, as well as allowing the model to distinguish between a measured zero value in $\vx$ and a missing value. 

\paragraph{Event Time Input Representation}

\begin{figure}[t]
\begin{minipage}{.49\textwidth}
    \centering
    \begin{adjustbox}{width=\textwidth}
\begin{tikzpicture}[scale=1.2,
        x={(1.0cm,0.0cm)}, y={(0.0cm,1.0cm), z={(-0.5cm,-0.1cm)}}% All grids are ok
        ]
        \begin{scope}[canvas is yz plane at x=4]
                \fill[fill=black!20] (0,-2) rectangle (3,0);
                \draw [step=0.5cm] (0,-2) grid (3,0);
        \end{scope}
        \begin{scope}[canvas is xz plane at y=3]
                \fill[fill=black!20] (0,-2) rectangle (4,0);
                \draw [step=0.5cm] (0,-2) grid (4,0);
        \end{scope}

        \fill[fill=red!30] (0,0) rectangle (3.5,0.5);
        \fill[fill=green!30] (3.5,0) rectangle (4,3);
        \fill[fill=yellow!30] (0,0.5) rectangle (0.5,3);
        \fill[fill=blue!30] (0.5,0.5) rectangle (1,3);
        \fill[fill=magenta!30] (1,0.5) rectangle (1.5,3);
        \fill[fill=cyan!30] (1.5,0.5) rectangle (2,3);
        \fill[fill=green!20] (2,0.5) rectangle (2.5,3);
        \fill[fill=black!20] (2.5,0.5) rectangle (3,3);
        \fill[fill=orange!30] (3,0.5) rectangle (3.5,3);
        \draw[step=0.5cm] (0,0) grid (4,3);

        \draw[ultra thick] (0,0) rectangle (3.5,0.5);
        \draw[ultra thick] (3.5,0) rectangle (4,3);
        \draw[ultra thick] (0,0.5) rectangle (3.5,3);

        % \node[align=center] at (-1,2.5) (events-label) {Time series \\ events};
        % \draw[thick] (events-label) -- (0,1.75);
        \node[align=center] at (5.8,2.5) (events-label) {Time series \\ events};
        \draw[thick] (events-label.west) -- (3.5,1.75);
        \node[align=center] at (-1.4,0.8) (static-label) {Static variable \\ embeddings};
        \draw[thick, shorten <=-0.2cm] (static-label) -- (0,0.25);
        \node[align=center] at (5.8,1.5) (rep-label) {Learned \\ {[REP]} token};
        \draw[thick] (rep-label) -- (4,1.25);

        % \draw[stealth] (0.5,-0.5) -- (3.5,-0.5);
        \node (time-label) at (2,-0.45) {Time dimension: length \(n_t+1\)};
        \draw[-stealth] ([xshift=-1.5cm,yshift=0.05cm] time-label.north) -- ([xshift=1.5cm,yshift=0.05cm] time-label.north);

        \draw[-stealth] (4.2,0) -- (4.8, 0.6);
        \node[align=center] at (5.5,-0.2) {Embedding \\ dimension: \\ length \(d\)};

        \draw[-stealth] (-0.12,2.8) -- (-0.12, 0.1);
        \node[align=center] at (-1.4,2.1) {Event dimension: \\ length \(n_e+1\)};
\end{tikzpicture}
\end{adjustbox}
    \caption{\small Structure of the processed input tensor. Event observations are binned across time into $n_t$ bins and mapped to $d$-dimensional embeddings, resulting in a $n_e \times n_t \times d$ tensor. Static variable embedding is concatenated to each time bin, and a learned [REP] token is appended to each event type including static dimension. This extends the tensor to $(n_e+1) \times (n_t+1) \times d$; the [REP] token output is used for downstream tasks.
    }
    \label{fig:input}
\end{minipage}
\hspace{0.02\textwidth}
\begin{minipage}{.49\textwidth}
    \begin{adjustbox}{width=0.84\textwidth}
\begin{tikzpicture}[scale=1.2,
        x={(1.0cm,0.0cm)}, y={(0.0cm,1.0cm), z={(-0.5cm,-0.1cm)}}% All grids are ok
        ]
        \begin{scope}[canvas is yz plane at x=4]
                \fill[fill=black!20] (0,-2) rectangle (3,0);
				\fill[fill=black!60] (1.5,-2) rectangle (2,0);
                \draw [step=0.5cm] (0,-2) grid (3,0);
        \end{scope}
        \begin{scope}[canvas is xz plane at y=3]
                \fill[fill=black!20] (0,-2) rectangle (4,0);
				\fill[fill=black!60] (1,-2) rectangle (1.5,0);
                \draw [step=0.5cm] (0,-2) grid (4,0);
        \end{scope}

        \fill[fill=red!30] (0,0) rectangle (3.5,0.5);
        \fill[fill=green!30] (3.5,0) rectangle (4,3);
        \fill[fill=yellow!30] (0,0.5) rectangle (0.5,3);
        \fill[fill=blue!30] (0.5,0.5) rectangle (1,3);
        \fill[fill=magenta!30] (1,0.5) rectangle (1.5,3);
        \fill[fill=cyan!30] (1.5,0.5) rectangle (2,3);
        \fill[fill=green!20] (2,0.5) rectangle (2.5,3);
        \fill[fill=black!20] (2.5,0.5) rectangle (3,3);
        \fill[fill=orange!30] (3,0.5) rectangle (3.5,3);
        \fill[fill=black!60] (1,0) rectangle (1.5,3);
        \fill[fill=black!60] (0,1.5) rectangle (4,2);
        \draw[step=0.5cm] (0,0) grid (4,3);

        \draw[ultra thick] (0,0) rectangle (3.5,0.5);
        \draw[ultra thick] (3.5,0) rectangle (4,3);
        \draw[ultra thick] (0,0.5) rectangle (3.5,3);

		\node at (2.5, 4.2) (z) {\(\bm{Z}\)};
		\draw[thick] (z) -- (2.5,3.77);

		\draw [decorate, decoration = {brace,mirror}, very thick] (1,-0.1) -- (1.5, -0.1);
		\draw[thick] (1.25, -0.17) -- (1.25,-0.4);
		\draw[thick] (1.25, -0.4) -- (0.3,-0.4);
		\draw[thick] (1.25, -0.4) -- (2.2,-0.4);
		\node[rectangle, draw, fill=black!10, align=center, minimum height=1cm] at (0.3,-1) (tvh) {Time bin \\ value head};
		\node[rectangle, draw, fill=black!10, align=center, minimum height=1cm] at (2.2,-1) (tph) {Time bin \\ pres.\ head};
		\draw[thick] (0.3, -0.4) -- (tvh);
		\draw[thick] (2.2, -0.4) -- (tph);

		\node at (0.3,-1.9) (tvy) {\(\hat{\vy}^\text{value}\)};
		\draw[thick, -stealth] (tvh) -- (tvy);
		\node at (2.2,-1.9) (tpy) {\(\hat{\vy}^\text{pres}\)};
		\draw[thick, -stealth] (tph) -- (tpy);

		\draw [decorate, decoration = {brace}, very thick] (-0.1,1.5) -- (-0.1,2);
		\draw[thick] (-0.13,1.75) -- (-1.65,1.75);
		\draw[thick] (-1.65,1.75) -- (-1.65,1.4);
		\draw[thick] (-1.65,1.4) -- (-2.85,1.4);
		\draw[thick] (-1.65,1.4) -- (-1,1.4);
		\node[rectangle, draw, fill=black!10, align=center, minimum height=1cm] at (-1,0.8) (eph) {Event type \\ pres.\ head};
		\node[rectangle, draw, fill=black!10, align=center, minimum height=1cm] at (-2.85,0.8) (evh) {Event type \\ value head};
		\draw[thick] (-2.85,1.4) -- (evh);
		\draw[thick] (-1,1.4) -- (eph);

		\node at (-2.85,-0.1) (evy) {\(\hat{\vy}^\text{value}\)};
		\draw[thick, -stealth] (evh) -- (evy);
		\node at (-0.8,-0.1) (epy) {\(\hat{\vy}^\text{pres}\)};
		\draw[thick, -stealth] (eph) -- (epy);

\end{tikzpicture}
\end{adjustbox}
    \caption{\small Diagram showing masking and output of SSL prediction heads. Randomly sampled slices along the time and event dimensions are replaced with a learned [MASK] embedding (shown in grey). Corresponding representations in the Transformer output $\bm{Z}$ are flattened and passed to MLP prediction heads to reconstruct the presence $\hat{\vy}^\text{pres}$ and value $\hat{\vy}^\text{value}$ of masked events.}
    \label{fig:ssl}

\end{minipage}
 \end{figure}

To construct the event time input representation to \modelabb{}, each event type \(i\) in time bin $j$ is mapped to a $d$-dimensional embedding. We use an MLP to map event value $x_{i,j}$ and corresponding count $m_{i,j}$ to an embedding $\bm{\phi}_{i,j} \in \R^d$: \(\bm{\phi}_{i,j} = \mbox{\bf MLP} \left( [x_{i,j}, \  \vp^m(m_{i,j})] \right)\),
where $[\cdot, \cdot]$ is the concatenation operation. Rather than directly concatenating the event value and count, which can lead to poor gradient scaling, we pass counts through an embedding function \(\vp^m(\cdot)\) that maps integer count values to discrete bins, then maps each bin to a learned scalar. Previous work on tabular data~\cite{gorishniy2021revisiting} has shown that learning scalars to represent counts is more effective than simply passing integer values, allowing the network to emphasize more salient differences in the numbers of observations and mitigate the effect of large outliers.

To incorporate static data, we embed all static variables into a \(d\)-dimensional vector using another MLP. This vector is concatenated to event embeddings in all time bins so each bin has access to static information: \(\bm{\phi}_{n_e+1,j} = \text{\bf MLP}(\vs), \, \forall j \in \{1,..., n_t\}\).
Incorporating the static variables into the input, rather than via late fusion as in recent work~\citep{tipirneni2022self,zhang2022graphguided}, enables our model to fully leverage such information in every layer, which is beneficial since static variables, such as age and sex, provide critical prior information that could considerably influence treatment strategies and outcomes.

Lastly, a learned $d$-dimensional patient representation token, [REP], is appended to the input for every event type: \(\bm{\phi}_{i,n_t+1} = \bm{[REP]}, \,\forall i \in \{1,..., n_e+1\}\).
The representation token aggregates relevant patient information, and the corresponding Transformer output is used for downstream classification tasks. 

The final input to our model is a three-dimensional tensor \(\bm{\Phi} \in \R^{(n_e+1) \times (n_t+1) \times d}\), where \(\bm{\Phi}_{i,j,\cdot}=\bm{\phi}_{i,j}\,\). Note that there is an extra dimension $n_e+1$ in the event representation to account for static input, and in the time sequence $n_t+1$ for the [REP] token. A diagram showing the structure of the full input tensor is given in Figure~\ref{fig:input}. This input representation allows the model to preserve the event/time information while binning adapts the length and granularity of input sequence, controlling model run-time complexity.  In comparison to input representations where each event is encoded separately~\citep{tipirneni2022self}, our approach reduces the Transformer layer computational complexity from \(O(n_p^2)\) to \(O(n_t^2 + n_e^2)\) in terms of time and memory, where generally, \(n_t,n_e \ll n_p\).

\subsection{DuETT Architecture}

\begin{figure}[t]
\centering
    \input{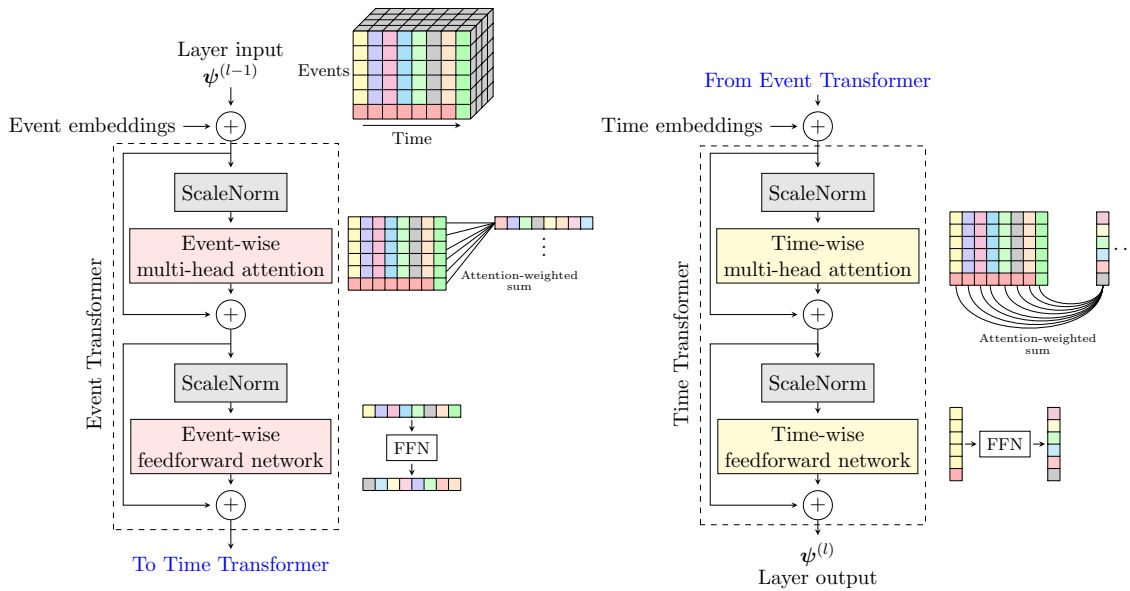}
    \vspace{-5mm}
    \caption{\small Diagram of a \layername{} layer. The layer receives a three-dimensional tensor with event type, time, and embedding dimensions. Event embeddings are added, then a Transformer sublayer operating across the event dimension is applied. Time embeddings are then added and a Transformer sublayer operating instead across the time dimension is applied. Side diagrams show the dimension on which Attention and FFN operations are applied. A pre-norm setup with ScaleNorm is used for a clean residual path across the entire network, preserving the input data structure. In the full model, multiple \layername{} layers are stacked together.
    }
    \label{fig:architecture}
\end{figure}

The overall structure of our \modelabb{} model is a series of \layername{} layers followed by classification or self-supervised learning heads. Each \layername{} layer is made up of two Transformer sublayers that attend along the event and time dimensions respectively. The first sublayer consists of multi-head attention over events followed by a feed-forward network operating along the event dimension, which can be collectively identified as an event transformer layer; the second sublayer consists of multi-head attention over time bins followed by a feed-forward network operating along the time dimension, the time transformer layer. The dual attention architecture enables our model to capture the two important modalities of EHR data, namely the types of events that are observed for a given patient and the times at which they are observed. Event-type and time bin embeddings are injected just before their respective sublayers. Embedding injections are done throughout the entire network, rather than just before the first layer, to ensure access and to emphasize the ordering information of data, especially in upper layers~\cite{gu2017non}. 

We denote the input tensor for the \(l\)-th \layername{} layer as $\vpsi^{(l-1)} \in \R^{(n_e+1) \times (n_t+1) \times d}$ with $\vpsi^{(0)} = \bm{\Phi}$. We use $\vpsi_{i,:,:} \in \R^{(n_t+1)d}$ and $\vpsi_{:,j,:} \in \R^{(n_e+1)d}$ to denote vectors flattened along the dimensions with colons. Considering each Transformer sublayer as a function that maps a sequence of inputs to a sequence of outputs, we can express the \layername{} layer \(l\) as an event $\text{Transformer}_{e}$ sublayer operating on a sequence of event type representations followed by a time $\text{Transformer}_{t}$ sublayer operating on a sequence of time bin representations:
\begin{equation}~\label{eq:tr}
	\begin{aligned}
		\vomega^{(l)}_{1,:,:}, \vomega^{(l)}_{2,:,:}, \vomega^{(l)}_{3,:,:},& \ldots = \\
		\text{Transformer}_{e}&\left(\vpsi^{(l-1)}_{1,:,:}\! +\vp^e_1,\ \vpsi^{(l-1)}_{2,:,:}\!+ \vp^e_2,\ \vpsi^{(l-1)}_{3,:,:}\! +\vp^e_3,\ \ldots\right) \\
		\vpsi^{(l)}_{:,1,:}, \vpsi^{(l)}_{:,2,:}, \vpsi^{(l)}_{:,3,:},& \ldots = \\
		\text{Transformer}_{t}&\left(\vomega^{(l)}_{:,1,:}\! +\vp^t_1,\ \vomega^{(l)}_{:,2,:}\! +\vp^t_2,\ \vomega^{(l)}_{:,3,:}\! +\vp^t_3,\ \ldots\right) \\
	\end{aligned}
\end{equation}
where \(\vp^e_i \in \R^{(n_t+1)d}\) is the event type embedding for the $i$'th event, and \(\vp^t_j \in \R^{(n_e+1)d}\) is the time embedding for the $j$'th time bin. After the two Transformer sub-layers the output is reshaped back into a 3D tensor $\vpsi^{(l)} \in \R^{(n_e+1) \times (n_t+1) \times d}$ and passed to the next layer. A diagram of this architecture is shown in Figure~\ref{fig:architecture}.

The internal architecture of the Transformer sublayers follows the original Transformer paper~\cite{vaswani2017attention} with two modifications: we use the now popular pre-LN setup as described in~\cite{xiong2020layer}, and use ScaleNorm instead of LayerNorm as in~\cite{nguyen2019transformers} to enchance training stability. Dropout is used on feed-forward and attention connections.

The event embeddings are learned separately for each event type since there is no inherent order to event types. For time bin embeddings, one approach is to use the positional encoding as in~\citep{vaswani2017attention}. However, since the overall length of time represented in each bin can vary from patient to patient, encoding only positions would discard potentially useful information about the time scale. To incorporate this information, our model learns embeddings calculated from the continuous time values representing each bin. We use the continuous value embedding (CVE) approach proposed in~\cite{tipirneni2022self}, which passes each time value through a fully connected feed-forward neural network with one hidden layer of size $\sqrt{(n_e+1)d}$ and a $\tanh$ activation, followed by an output layer that produces a time embedding in $\R^{(n_e+1)d}$. In addition to incorporating continuous time information, the neural network is able to learn an embedding function that is well adapted to the data. The time value for a given bin is calculated as the difference between the bin end time and start of the patient's stay, and represents the (fractional) number of days that have passed since the start of the stay.

The output of \modelabb{} is a representation tensor $\bm{Z} \in \R^{(n_e+1) \times (n_t+1) \times d}$. Event and time bin representations, $\bm{Z}_{i,:,:} \in R^{(n_t+1)d}$ and $\bm{Z}_{:,j,:}\in R^{(n_e+1)d}$ respectively, are used for self-supervised learning, while the [REP] token representation $\bm{Z}_{:,n_t+1,:}$ is used for supervised tasks as described in the following section.

\subsection{Training}

The model is trained in two phases: self-supervised pre-training followed by supervised fine-tuning on downstream tasks. 

\paragraph{SSL pre-training}

During pre-training, we aim to train the model to capture important clinical priors. We therefore select tasks that capture useful information about the underlying patient state using the observed data.

Masked event modelling predicts the values of masked inputs based on other inputs, and resulting in the modeling learning useful information about the clinical relationships between different observations. The input sparsity structure also reflects important aspects of the patient's condition, with missing values providing information about the clinician's intent to treat or measure the event in question. To capture this, we design a self-supervised task based on predicting both the presence/absence of an event and its value. 

To capture relationships in both event and time dimensions for a more complete view of the patient state, we introduce a masking scheme along {\it both} the time and event dimensions. Time-wise masking encourages the model to learn how a measurement made at a certain time relates to patient state at different times, while event-wise masking focuses on how certain kinds of patient measurements relate to other kinds of measurements across all time bins. We find that using both value and presence losses across both event and time dimension produces a rich clinical prior with improved performance compared to simpler masking and loss schemes, as we show in Section~\ref{sec:discussion}.

The masking is done by replacing selected inputs with a learned embedding [MASK] $\in \R^d$. For event-wise masking we randomly select a set of event types to mask across all time steps, e.g. for a selected event type \(i\), all inputs \(\bm{\phi}_{i,1}, \bm{\phi}_{i,2}, \ldots, \bm{\phi}_{i,n_t}\) are replaced with [MASK]. Similarly for time-wise masking, we select a set of time bins to mask across all event times, such that for a selected time bin \(j\), all inputs \(\bm{\phi}_{1,j}, \bm{\phi}_{2,j}, \ldots, \bm{\phi}_{n_e,j}\) are replaced with [MASK].
The number of time bins and event types to mask at each training step is set as a hyperparameter, and they are sampled uniformly. 
The final Transformer outputs $\bm{Z}_{i,:,:}$ and $\bm{Z}_{:,j,:}$, corresponding to the masked input event type $i$ and time bin $j$ respectively, are passed to presence and value MLP prediction heads to produce the predictions $\hat{\vy}^\text{pres}$ and $\hat{\vy}^\text{value}$. The presence head performs binary classification, predicting whether target events were observed in the given time bins, and the value head predicts the corresponding event value. These predictions are then compared with the actual presence and values using cross entropy and squared error losses respectively. Different heads are trained for the time and event dimensions. A diagram illustrating the time and event-wise masking and prediction tasks is shown in Figure~\ref{fig:ssl}.

For a single masked input at \((i,j)\), the pre-training loss is given by:
\begin{align}
\label{eqn:SSLloss}
	&\gL_{i,j} = \gL_{i,j}^\text{value} + \alpha\gL_{i,j}^\text{pres} \nonumber \\
    &\gL_{i,j}^\text{value} = \ind[m_{i,j} > 0]\left(\hat{y}^\text{value}_{i,j} - x_{i,j}\right)^2  \\
    &\gL_{i,j}^\text{pres} = - \ind[m_{i,j} > 0]\log(\hat{y}^{\text{pres}}_{i,j}) - \ind[m_{i,j} = 0]\log(1-\hat{y}^{\text{pres}}_{i,j}) \nonumber
\end{align}
where $\alpha$ is a hyperparameter that controls the contribution of each task and $\ind$ is an indicator function. For each masked time bin or event type, we average the loss across all relevant masked inputs.

\paragraph{Fine-tuning}
During fine-tuning, we use the patient representation $\bm{Z}_{:,n_t+1,:}$ produced by the Transformer from the [REP] input, and attach heads tailored to the downstream tasks. Note that the [REP] embedding value is learned at this stage, since its output is not used during pre-training. For the tasks explored in this paper we use MLP classification heads with sigmoid output and binary cross entropy loss.

\section{Cohort} \label{experiment}

We evaluate our proposed model on two widely used EHR datasets: MIMIC-IV \citep{mimic_iv_2022} and the PhysioNet/CinC Challenge 2012 \citep{silva_2012}. In this section, we present our data preprocessing steps, experimental designs, and model performances compared to the leading baselines. We also present experiments to demonstrate the quality of the learned representations and conduct an ablation study to evaluate the impact of the components of our approach. We consider the following tasks to evaluate our models: 

\textbf{MIMIC-IV} \citep{mimic_iv_2022} is a public dataset that contains retrospective, deidentified data of patients admitted to the ICU or the emergency department (ED) at the Beth Israel Deaconess Medical Center between 2008 and 2019. This dataset contains data of various modalities: time series data, static tabular data, and medical images.
We evaluate tasks on a derived ICU dataset, containing $53\,150$ patients with $69\,211$ admissions, and an ED dataset, containing $112\,577$ patients with $213\,911$ admissions. 
For both datasets, we use a patient-level 70\%:15\%:15\% split between the training, validation, and test sets.

For the ICU dataset, we follow~\cite{nature_benchmark_2019} in defining mortality prediction and phenotype classification tasks. We exclude patients below 18 years of age and patients with no chart or lab events recorded during the stay. Unlike \cite{nature_benchmark_2019}, we do not exclude patients with multiple ICU stays or transfers between ICU units during their stay. This results in a larger dataset that more closely mimics the practical use of a machine learning system in a hospital setting. The mortality prediction task uses the first $48$ hours of the patient stay as the input time window, predicting whether death occurs later during the hospital stay and has 13\% positive instances. Patients with stays of less than $48$ hours and patients with no recorded events before $48$ hours are excluded from this task. The phenotype classification task uses the entire ICU stay as the input time window and uses a multi-label classification target, predicting 25 common hospital diagnoses. Details are provided in Appendix~\ref{mimic-appendix}.
We include all input variables used in~\cite{nature_benchmark_2019} as well as a number of static variables and all chart and lab events that are observed in more than $50\%$ of ICU stays. This substantially increases the set of variables, and provides a rich input signal to the model. The variables are listed in Appendix~\ref{mimic-appendix}. 

For the ED dataset, we define a task of predicting whether a patient will be transferred to the ICU during their stay, with a target positive rate of 9\%. Our feature window is the first six hours of the ED stay, and patients with an ED stay shorter than six hours are excluded. We again exclude patients below 18 years of age. We also exclude patients that are not formally admitted to the hospital, since these stays are generally very short and have little data available.

\textbf{PhysioNet-2012}~\citep{silva_2012} is a standardized dataset with the task of predicting in-hospital mortality after the first $48$ hours of patient stays in the ICU, where $14\%$ of mortality labels are positive. The dataset consists of $12,000$ ICU stays with $42$ different variables including $37$ time series event-types. The details of the dataset, including data statistics, are provided in \cite{silva_2012}. We use the torchtime~\citep{darke2022torchtime} data library to preprocess the data in a standard way and to split the dataset into training, validation and test sets $(70\%:15\%:15\%)$.

\section{Experiments}

\paragraph{Preprocessing and hyperparameters} We apply zero-mean and unit-standard deviation normalization to all inputs in $\vx$. We also clip outliers using a threshold of three median absolute deviations from the median. These steps allow for stable training without prior domain knowledge of all normal variable ranges. Binary cross entropy loss is used for all supervised training. We also weight positive and negative instances according to the target positive fraction so that they receive equal weight in the loss.

We provide full hyperparameter settings for \Modelabb{} in Appendix~\ref{our-hypers-appendix} and our code repository. To aggregate events in each time bin, we take the last observed value of each. We perform self-supervised pre-training for 300 epochs using AdamW~\citep{loshchilov2017decoupled}. The learning rate is scheduled to have linear warmup followed by inverse square-root decay, as in typical Transformer training. One time step and one event type per iteration are masked out for self-supervised learning tasks, as masking more steps did not improve performance. After pre-training, we use the weights from the epoch with lowest validation loss for fine-tuning.

We fine-tune \modelabb{} for 30 epochs for MIMIC-IV and 50 epochs for PhysioNet. We average weights from the five epochs with the best performance on the validation set to produce our final model. We use the same architecture across all datasets and tasks, only varying a small number of optimizer and regularizer settings, showing that the architecture generalizes well without extensive tuning.

We run all experiments on a single NVidia A6000 GPU. The most resource-intensive \modelabb{} pre-training and fine-tuning procedure only uses 7GB of GPU memory and completes within two days. 

To ensure reproducibility, we will publish implementation code for our model as well as the IDs for patient-level splits on our Github page along with the camera-ready version of the paper. We generated all results that have reported standard deviations using consistent random seeds from $2020-2022$. All other experiments used a random seed of $2020$.

\begin{table*}[t]
	\caption{\small Performance on tasks across MIMIC-IV and PhysioNet-2012 datasets. Results show mean and standard deviation over three fixed seeds. Phenotyping metrics are macro-averaged. * and ** represent p-value significance relative to the next best model. *: \(p<0.05\), **: \(p<0.001\)}
	\label{t:mimic-performance}
 \vspace{-5mm}
    	\begin{center}
    	    \resizebox{0.92\textwidth}{!}{
    			\begin{tabular}{l|ll|ll|}
    				\toprule
    				\multirow{3}{*}{\bf Model} &
    				\multicolumn{4}{c|}{\bf MIMIC-IV ICU} \\
    				\cmidrule{2-5}
    				& 
    				\multicolumn{2}{c|}{\bf Mortality} &
    				\multicolumn{2}{c|}{\bf Phenotyping}\\
    				\cmidrule{2-5}
    				& 
    				{\bf ROC-AUC} &
    				{\bf PR-AUC} &
    				{\bf ROC-AUC} &
    				{\bf PR-AUC} \\
    				\midrule
    				XGBoost & 0.886\tpm 0.003 &  0.593\tpm 0.004 & 0.829\tpm 0.001 &  0.589\tpm 0.001  \\
    				LSTM    &  0.881\tpm 0.001 & 0.533\tpm 0.006 & 0.756\tpm 0.002 & 0.447\tpm 0.001  \\
    				mTAND & 0.864\tpm 0.002 & 0.540\tpm 0.007 & 0.812\tpm 0.001 & 0.553\tpm 0.003  \\
    				Raindrop & 0.878\tpm 0.001 & 0.546\tpm 0.002 & 0.824\tpm 0.001 & 0.577\tpm 0.003 \\
                        STraTS  &  0.882\tpm0.004  & 0.552\tpm 0.013 & 0.820\tpm0.001 & 0.565\tpm0.002  \\
    				\midrule
    				DuETT (Ours) & \bf 0.912\tpm 0.02* & \bf0.627\tpm 0.002** & \bf 0.838\tpm 0.001** &\bf  0.604\tpm 0.002** \\
    				\bottomrule
    			\end{tabular}}
       
       \vspace{3mm}
       
       \resizebox{0.92\textwidth}{!}{
    			\begin{tabular}{l|ll|ll|}
    				\toprule
    				\multirow{3}{*}{\bf Model} &
    				\multicolumn{2}{c|}{\bf MIMIC-IV ED} &
    				\multicolumn{2}{c|}{\bf PhysioNet-2012} \\
    				\cmidrule{2-5}
    				& 
    				\multicolumn{2}{c|}{\bf Transfer to ICU} &
    				\multicolumn{2}{c|}{\bf Mortality} \\
    				\cmidrule{2-5}
    				& 
    				{\bf ROC-AUC} &
    				{\bf PR-AUC} &
    				{\bf ROC-AUC} &
    				{\bf PR-AUC} \\
    				\midrule
    				XGBoost & 0.833\tpm 0.0001 &  0.446\tpm 0.001 & 0.865\tpm0.001 & 0.531\tpm0.009 \\
    				LSTM   & 0.777\tpm0.06  & 0.327\tpm0.1  & 0.848\tpm 0.002 & 0.494\tpm 0.002  \\
    				mTAND & 0.807\tpm0.001  & 0.398\tpm0.005  & 0.857\tpm 0.001 & 0.515\tpm 0.007\\
    				Raindrop & 0.821\tpm0.001  & 0.413\tpm0.004  &  0.838\tpm 0.009 & 0.479\tpm 0.002 \\
                        STraTS   & 0.789\tpm0.01  & 0.329\tpm0.03  & 0.852\tpm 0.008  &  0.527\tpm 0.006  \\
    				\midrule
    				DuETT (Ours) & \bf 0.841\tpm 0.0007** &\bf  0.467\tpm 0.002** & \bf 0.872\tpm 0.001**  & \bf 0.564\tpm 0.003**  \\
    				\bottomrule
    			\end{tabular}}
    	\end{center}
    \vspace{-3mm}
\end{table*}

We show that \modelabb{} outperforms a range of baseline models, including the well established XGBoost and LSTM baselines as well as state-of-the-art deep learning models:
\begin{itemize}[nosep,leftmargin=*]
    \item \textbf{XGBoost} \citep{chen2016xgboost}:  A scalable tree-based gradient boosting model that has been shown to outperform deep learning models on tabular data \citep{shwartz2022tabular}.
    \item \textbf{LSTM} \citep{graves2012long}: A standard time series RNN. We use the same binned input format as for \modelabb{}.
    \item \textbf{mTAND} \cite{shukla2021multitime}: An encoder-decoder based model that uses an attention module to interpolate irregular and sparse multivariate time series. It uses an unsupervised training task.
    \item \textbf{STraTS} \citep{tipirneni2022self}: A Transformer-based model where every observation is embedded separately to produce the Transformer input sequence. It uses a self-supervised pre-training approach.
    \item \textbf{Raindrop} \cite{zhang2022graphguided}: A graph-based neural network model that uses message passing between time series variables to learn relevant relationships.
\end{itemize}

\subsection{Quantitative Results} 
We highlight our results in Table~\ref{t:mimic-performance} and the details on baseline implementations are given in Appendix \ref{baseline-hypers-appendix}. To provide a fair comparison, we ensure that all static variables as well as time series variables are provided to the baseline models. We report the mean and standard deviation of ROC-AUC and PR-AUC over three supervised training runs using different random seeds. We note that while ROC-AUC and PR-AUC demonstrate similar trends, PR-AUC provides more discrimination between methods. 

It is worthwhile to first note that a tuned XGBoost model is one of the strongest baseline across both datasets on all tasks, outperforming prior neural architectures on this task. This observation is in agreement with previous work that investigated behavior of tree-based models on tabular datasets ~\cite{grinsztajn2022tree}. The superior and consistent performance indicates that XGBoost, with appropriate feature engineering and hyperparameter tuning, is still very competitive with neural network models for sparse irregular time series, and should be included in evaluation of future methods. Over a well tuned XGBoost baseline, \ModelAbb{} significantly outperforms all baselines across all datasets and tasks.

\subsection{Representation Quality} \label{sec:rep-quality}

\begin{figure}[t]
\begin{minipage}{.49\textwidth}
\resizebox{\textwidth}{!}{\begin{tikzpicture}[font=\normalsize]
  \begin{axis}[
        ybar, axis on top,
        height=7cm, width=12cm,
        bar width=0.9cm,
        ymajorgrids, tick align=inside,
        enlarge y limits={value=.1,upper},
        ymin=0.4, ymax=1,
        axis x line*=bottom,
        axis y line*=left,
        y axis line style={opacity=0},
        tickwidth=0pt,
		enlarge x limits=0.5,
        legend style={
            at={(0.5,-0.13)},
            anchor=north,
            legend columns=-1,
            /tikz/every even column/.append style={column sep=0.5cm}
        },
		area legend,
        symbolic x coords={
		ROC-AUC, PR-AUC},
       xtick=data,
       nodes near coords={
		   \pgfmathprintnumber[precision=3]{\pgfplotspointmeta}
       }
    ]
    \addplot [draw=none, fill=green!50] coordinates {
      (ROC-AUC, 0.847)
	  (PR-AUC, 0.473) };
   \addplot [draw=none,fill=teal!50] coordinates {
      (ROC-AUC, 0.875)
	  (PR-AUC, 0.525) };
   \addplot [draw=none, fill=red!60] coordinates {
      (ROC-AUC, 0.888)
	  (PR-AUC, 0.587) };
    \legend{mTAND, STraTS, DuETT}
  \end{axis}
  \end{tikzpicture}}
     \caption{Performance on the MIMIC-IV mortality prediction task with pre-trained encoders where the encoder weights are frozen during supervised fine-tuning.}
     \label{fig:frozen-encoder}
\end{minipage}
\hspace{0.02\textwidth}
\begin{minipage}{.49\textwidth}
    \pgfplotsset{
        width=\linewidth,
        title={},
        xlabel={\% of labelled training data},
    	xtick = {0,1,2,3,4,5,6,7,8,9,10},
        xticklabels={0,10,20,30,40,50,60,70,80,90,100},
    	legend pos=outer north east,}
    \centering
\resizebox{\columnwidth}{!}{
\begin{tikzpicture}[font=\Large]

    \begin{axis}[
        ylabel={PR-AUC},
        legend style = { column sep = 10pt, legend columns = -1, legend to name = grouplegend,},
        scale only axis,
        width=9cm,
        height=5cm,
    	ymin=0.36,
     legend columns=3
     ]
    \addplot[ % TESS Final
        color=red,
        mark=*,
        ]
        coordinates {
    		(0.5,0.4710)
    		(1,0.5204)
    		(2,0.5529)
    		(5,0.6022)
    		(10,0.6274)
        }; \addlegendentry{\modelabb}
    \addplot[ % LSTM Final
        color=blue,
        mark=triangle*,
        ]
        coordinates {
    		(0.5,0.3783)
    		(1,0.4563)
    		(2,0.4693)
    %		(4,0.5074)
     		(5,0.5077)
    	%	(6,0.5178)
    % 		(8,0.5349)
    		(10,0.5332)
        }; \addlegendentry{LSTM}
    \addplot[ % STraTS Final
        color=teal,
        mark=square*,
        mark options={scale=0.5}
        ]
        coordinates {
    		(0.5,0.4634)
    		(1,0.5059)
    		(2,0.5204)
    % 		(4,0.5212)
    		(5,0.54411)
    % 		(6,0.5433)
    % 		(8,0.5392)
    		(10,0.5519)
        }; \addlegendentry{STraTS}
    \addplot[ % mTAN Final
        color=green,
        mark=diamond*,
        ]
        coordinates {
    		(0.5,0.4445)
    		(1,0.4630)
    		(2,0.4867)
    % 		(4,0.5069)
            (5,0.5236)
            % (6,0.5219)
            % (8,0.5284)
    		(10,0.5399)
        }; \addlegendentry{mTAND}
     \addplot[ % XGB Final
        color=orange,
        mark=pentagon*,
        ]
        coordinates {
    		(0.5,0.4019)
    		(1,0.4755)
    		(2,0.5176)
    % 		(4,0.5458)
    		(5,0.5645)
    % 		(6,0.5688)
    % 		(8,0.5789)
    		(10,0.593)
        }; \addlegendentry{XGBoost}
    \addplot[ % Raindrop Final
        color=cyan,
        mark=pentagon*,
        ]
        coordinates {
    		(0.5,0.2008)
    		(1,0.4349)
    		(2,0.4780)
    % 		(4,0.5157)
    		(5,0.5197)
    % 		(6,0.5219)
    % 		(8,0.5331)
    		(10,0.5456)
        }; \addlegendentry{Raindrop}
    \end{axis}
    \node at (4.5,-2.0) {\pgfplotslegendfromname{grouplegend}}; 
\end{tikzpicture}}
    \captionof{figure}{Performance on MIMIC-IV mortality prediction task using different percentages of labelled data.}
    \label{fig:percentage_labeled}
\end{minipage}
 \end{figure}

To evaluate the quality of the representations learned by our model compared to other baselines, we first carry out self-supervised pre-training, then freeze the encoder weights and fine-tune the model with a linear classifier attached to the encoder. Among our baselines, mTAND and STraTS can also be trained in this way. The original mTAND model augments supervised training with an unsupervised component, so for a fair comparison, we pre-train the mTAND encoder-decoder architecture using only the unsupervised loss. As shown in Figure~\ref{fig:frozen-encoder}, \modelabb{} outperforms both of these baselines. This demonstrates the ability of \modelabb{} to learn useful patient representations from our self-supervised pre-training approach, without relying on labelled data. We also see that results for all models are lower than SSL combined with end-to-end fine-tuning  in Table~\ref{t:mimic-performance}, indicating that it is preferable to fine-tune all weights.

Next, we study the performance of \modelabb{} when only a fraction of labelled data is used for supervised fine-tuning. We highlight two key findings. First, Figure~\ref{fig:percentage_labeled} shows that \modelabb{} outperforms the baselines consistently across all fractions of labelled data.
Second, the performance gap relative to the self-supervised Transformer baseline, STraTS, widens with more labelled data, while the gain of \modelabb{} over XGBoost increases as the percentage of labelled data decreases, demonstrating the effectiveness of SSL with sparse labels.

\begin{figure*}[t]
    \centering
    \resizebox{\textwidth}{!}{
    \includegraphics{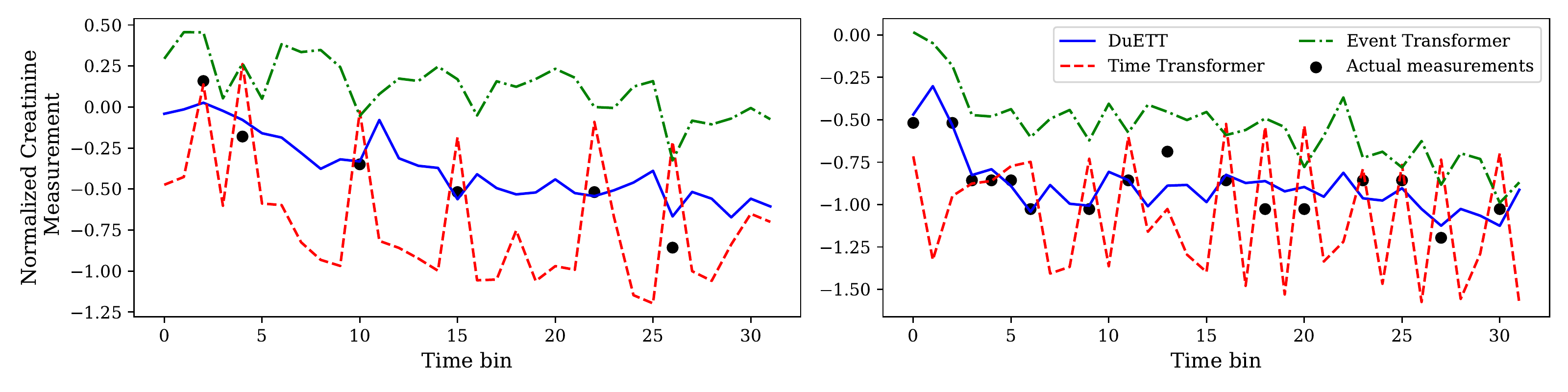}}
    \caption{\small     Reconstruction of masked creatinine measurements for two random validation set patients in the MIMIC-IV ICU dataset. DuETT is compared to models of the same depth that use only event Transformer sublayers or only time Transformer sublayers. The overall validation set reconstruction mean squared error (MSE) follows the same trend: \modelabb{} has a masked event MSE of 0.0754 while the time Transformer has MSE of 0.0867 and the event Transformer has MSE of 0.0916.}
    \label{fig:creatinine}
\end{figure*}

Finally, we show an example of the model capabilities learned during pre-training in Figure~\ref{fig:creatinine}. We mask all serum creatinine variables and use the pre-trained model to reconstruct the creatinine levels of two sample patients. Creatinine is an important marker of kidney function that is commonly measured in ICUs. \modelabb{} successfully reconstructs trends in the value over time, whereas model variants that only use event Transformer or time Transformer sublayers show substantial errors. This is a task that requires sophisticated modelling of the relationship between creatinine and other observed event types as well as the evolution of values over time, and DuETT's modelling of event and time dimensions is well-adapted for this task.

\section{Discussion}\label{sec:discussion}

Our results demonstrate that \modelabb{} can outperform existing state-of-the-art methods, which has obvious benefits for clinical use, and with little hyperparameter tuning required between different tasks. \modelabb{} also shows good performance in generating patient representations and training with limited labelled data, which is a common constraint in practical health care models. A practical application of this capability is pretraining one large model with as wide a range of hospital data as is available and then fine-tuning for downstream tasks as needed. An example of this is the shared pretraining between the MIMIC-IV ICU mortality and phenotyping tasks. Patient representations generated by \modelabb{} can also be used as inputs to other machine learning models.

We argue that our performance gains are driven by the ability of \layername{} layers to attend across both time and event dimensions, with our self-supervised learning and input representation design decisions also being critical. To support this claim, we conduct an extensive ablation study to evaluate the importance of key components of our approach, with results shown in Table~\ref{tab:ablation-studys}. 
All ablations measure the effect of only making the specified change to \modelabb{} in isolation. We discuss each category of results in turn below. 

\begin{table}[t]
    \caption{\small Ablation study on the MIMIC-IV mortality task, measuring the impact of making the specified change to \modelabb{}. The $\Delta$ column gives the difference in PR-AUC from \modelabb{}. }
    \centering
    \small
    \begin{tabular}{llcc}
    \toprule
    \multicolumn{2}{l}{\textbf{Experiment}} & \textbf{PR-AUC} & \textbf{$\Delta$} \\
          \midrule
    \multicolumn{2}{l}{\textbf{\Modelabb{}}} & \bf 0.627\tpm 0.002 & --\\
    \midrule
    \ \ \parbox[t]{2mm}{\multirow{2}{*}{\rotatebox[origin=c]{90}{\textbf{\scriptsize Attn.}}}} 
        & Event Transformer only & 0.609\tpm 0.003 & -0.018\\
    \ \ & Time Transformer only & 0.587\tpm 0.004 & -0.040\\
    \midrule
    \ \ \parbox[t]{2mm}{\multirow{5}{*}{\rotatebox[origin=c]{90}{\textbf{\scriptsize SSL}}}} 
        & Value loss only & 0.611\tpm 0.005 & -0.016 \\
 
    \ \ & Presence loss only & 0.593\tpm 0.003  & -0.034 \\
    \ \ & Time bin masking only & 0.612\tpm 0.007 & -0.015 \\
    \ \ & Event type masking only & 0.577\tpm 0.001 & -0.050 \\
    \ \ & No SSL & 0.556\tpm 0.003 & -0.071 \\
    \midrule
    
    \ \parbox[t]{2mm}{\multirow{4}{*}{\rotatebox[origin=c]{90}{\textbf{\scriptsize Input}}}} \ 
        & Binning with mean aggregation & 0.618\tpm 0.001 & -0.009 \\
    \ \ & Binning with max aggregation & 0.616\tpm 0.003 & -0.011 \\
    \ \ & First layer embedding only & 0.615\tpm 0.004 & -0.012 \\
    \ \ & Late static input fusion & 0.610\tpm 0.003 & -0.017  \\
    \bottomrule
    \end{tabular}
    \label{tab:ablation-studys}
\end{table}

\paragraph{Event and Time Transformer Ablation}

To investigate the impact of using \modelabb{} layers, we ran ablation experiments by substituting all Transformer sublayers with only a single type, either an event Transformer or a time Transformer. As shown in Table~\ref{tab:ablation-studys}, both substitutions result in significant decreases in performance, indicating that the dual event time transformer structure is essential to the performance of \modelabb{}. It is interesting to note that the model with only event Transformer layers performs better ($0.022$) than the time Transformer only model, while past applications~\cite{tipirneni2022self} of Transformers on time series mainly focus on applying attention along the time dimension. The performance of the event Transformer only model, without attention over time bins, suggests that in multivariate time series datasets, relationships between different input/event types is just as useful and important as the relationships between neighbouring time steps; this is naturally handled by DuETT's dual event time Transformer layer.

\paragraph{Self-Supervised Learning Ablation}

For the "No SSL" ablation in Table~\ref{tab:ablation-studys}, we skip the pre-training phase and directly train our model on the labelled data in a supervised manner. This leads to a $0.071$ drop in PR-AUC, showing that self-supervised pre-training is an essential component in the superior performance of \modelabb{}. The "Value/Presence loss only" ablations are done by omitting the corresponding loss term in Equation \ref{eqn:SSLloss} during pre-training. Pre-training with only presence loss (no value prediction) decreases the PR-AUC by $0.034$, while pre-training with only value loss (omitting presence prediction) leads to a smaller drop of $0.015$. This gap demonstrates that numerical event values contain more information than the presence/absence of events, but both results are significantly lower than training with the full loss, suggesting that both losses are important components of \modelabb{}. We also ablate masking strategies, where "time bin / event type masking only" corresponds to masking and reconstructing results only along one or the other dimension; this can be visualized as having either the horizontal or the vertical masking in Figure \ref{fig:ssl}, but not both. The results again show a drop in performance for both of these configurations.

\paragraph{Input Representation Ablation}

For ablations on the input representation, we first investigate using different aggregation functions in each time bin.
Using maximum or mean value aggregation showed a small but observable drop in PR-AUC compared to using the last observed value. This suggests that our choice of aggregation function is most suitable for the current EHR tasks and datasets, but there is flexibility in choosing task-specific or dataset-specific aggregation functions. Ablation on only injecting event type embeddings $\vp^e_i$ and time bin embeddings $\vp^t_j$ (see Equation~\ref{eq:tr}) at the first layer also causes a noticeable drop, which supports our decision of injecting temporal and event type information at each layer via the time and event embedding.
Performing late static input fusion, meaning not providing static variables at the input layer, but only at the classification head, leads to a $0.017$ decrease in PR-AUC. This demonstrates the importance of providing the \layername{} layers access to this important patient background information.

\paragraph{Limitations}

Our study is limited to EHR data from a single hospital stay rather than EHR tasks where information is accumulated across multiple encounters, due to a lack of available detailed datasets. In its current form, the proposed model does not directly incorporate text or imaging data, though incorporating multi-modal information into this model is in line with our future research direction.

\section{Conclusion}

We introduce \modelabb{}, a Dual Event Time Transformer model that attends and processes events across both semantic dimensions of multivariate time series data. We build a self-supervised model for hospital EHR data, along with appropriate input processing and self-supervised learning tasks. Our experiments show that this architecture outperforms state-of-the-art models across a number of tasks, and is especially effective in learning useful information during self-supervised pre-training. We believe the ability of the \modelabb{} architecture to naturally process information in event and time dimensions makes it a robust model for multivariate time series modelling problems in general. For future work, we would like to apply our approach to other health care data and to sparse irregular time-series data in domains beyond health care. We believe that advancing the state of the art in self-supervised Transformer-based models will help drive substantial improvements in future health care modelling.  

\bibliography{main}

\clearpage
\appendix
\section{\ModelAbb{} Architecture Details} \label{our-hypers-appendix}

We use the following dimensions for the subnetworks within \modelabb{}. Value and presence prediction heads are both a single linear layer. We also use a linear layer for the input embedding MLP. The observation count embedding \(\vp^m\) uses distinct bins for each integer from 0 to 14, and another bin for counts \(\geq 15\). Our model uses 2 \layername{} layers, with a total of 4 Transformer sublayers. The Transformers have an internal feedforward dimension of 512. The classification head has one hidden layer of size 64 and batch normalization after the hidden layer. The static data encoder has one hidden layer of size 128 and batch normalization after the hidden layer. We use $n_t=32$ time steps.

Complete hyperparameter specifications are given in our code repository: \url{https://github.com/layer6ai-labs/DuETT}.

\section{Baseline Details}
\label{baseline-hypers-appendix}

For XGBoost, we use the same aggregated input representation as for \modelabb, with all \(\vx\), \(\vm\), and \(\vs\) values concatenated into a feature vector. However, we find that XGBoost does not handle the sparsity of inputs well, and so we impute missing \(\vx\) values using the last previously observed value when available. We perform random tuning with 100 tests using the hyperparameter distributions given in Table \ref{t:xgb-hypers} and use the configuration with best PR-AUC on the validation set.

\begin{table}[t]
	\caption{XGBoost Hyperparameter Tuning Distributions}
	\label{t:xgb-hypers}
	\begin{center}
		\begin{tabular}{ll}
			\toprule
			\textbf{Hyperparameter} & \textbf{Distribution}  \\
			\midrule
			Number of rounds & Uniform on \(\{50,51,\hdots,250\}\) \\
			\texttt{max\_depth} & Uniform on \(\{2,3,\hdots,16\}\) \\
			\texttt{eta} & Log-uniform on \([0.001,1]\) \\
			\texttt{lambda} & Log-uniform on \([0.001,1]\) \\
			\texttt{alpha} & Log-uniform on \([0.001,1]\) \\
			\texttt{subsample} & Uniform on \([0.2,1]\) \\
			\texttt{min\_child\_weight} & Log-uniform on \([0.01,100]\)
		\end{tabular}
	\end{center}
\end{table}

For mTAND, we use the configuration/hyperparameters given in \cite{shukla2021multitime} and their published code repository, using their PhysioNet hyperparameters for our PhysioNet tests and their MIMIC-III hyperparameters for our MIMIC-IV tests. Unlike the datasets evaluated in their paper, our MIMIC-IV phenotyping task uses arbitrarily long patient stays as input data, making it infeasible to train with the provided configurations. To mitigate this issue, for phenotyping only, we increase the quantization windows from 5 to 30 minutes and we limit the length of input data to the first two weeks of the patient stay. We also find that our zero-mean unit-variance normalization massively increases the mTAND reconstruction loss and reduces performance. For all tasks, we instead scale all variables to range from 0 to 1, matching their provided code. Further, we encode the static variables as time series with one sample as an input, which matches the mTAN code repository.

For Raindrop, we use the configuration/hyperparameters given in \cite{zhang2022graphguided} and their published code repository for PhysioNet-2012. For our PhysioNet tests, we use the raw time steps given in the dataset and do not discretize time. Unlike PhysioNet, the set of time steps at which observations can be made in MIMIC-IV is not limited, making it infeasible to use raw inputs. To provide a fair comparison on MIMIC-IV, we use the same discretized time bins as for \modelabb. The static data is passed directly into the Raindrop model as their implementation also handles the static data along with the time series data. 

For STraTS, the configuration/hyperparameters are set according \cite{tipirneni2022self} and their published code repository. As suggested in the paper, we set the maximum number of observations to the $99^{th}$ percentile of the observations in the $48h$ observation window. This results in $1832$ and $1898$ maximum sequence length for MIMIC-IV and PhysioNet respectively. The static data for STraTS is passed through a feed-forward neural network to obtain the embedding before concatenating with the time series embedding and passing through the final dense layer as described in the original paper.       

\section{MIMIC-IV Benchmark Details} \label{mimic-appendix}

We use all variables from \cite{nature_benchmark_2019} except GCS total, which does not have a corresponding item ID in MIMIC-IV, plus chart and lab variables observed in 50\% or more of patient stays. This amounts to $85$ chart event variables and $29$ lab event variables. For ICU tasks, we include $9$ static variables. For ED tasks, chart events are not available, but a subset of them are regularly recorded as vital signs. We also use ten patient and triage-related static variables for ED tasks. All variables are given in Table~\ref{t:mimic-variables}. The categorical variables are encoded using one-hot encoding.

\footnotesize{
\begin{longtable}[c]{llll}
\toprule
\textbf{Variable} & \textbf{Type} & \textbf{Source} & \textbf{Item IDs} \\
      \midrule
	  \multicolumn{4}{c}{\bf Variables from \cite{nature_benchmark_2019}} \\
	  \midrule
 Capillary refill rate & Time series & & 223951, 224308 \\
 Diastolic blood pressure & Time series & & 220051, 220180, \\
 & & & 224643, 225310, \\        & & & 227242 \\
 Fraction inspired oxygen & Time series & & 223835 \\
 Glasgow coma scale eye opening & Time series & & 220739 \\
 Glasgow coma scale verbal response & Time series & & 223900 \\
 Glasgow coma scale motor response & Time series & & 223901 \\
 Glucose & Time series & & 220621, 225664, \\ 
            &  &  & 226537, 228388 \\
 Heart rate & Time series & & 220045 \\
 Height & Time series & & 226707, 226730 \\
 Mean blood pressure & Time series & & 220052, 220181 \\
 Oxygen saturation & Time series & & 220227, 220277 \\
 Respiratory rate & Time series & & 220210, 223851, \\            & & & 224689, 224690 \\
 Systolic blood pressure & Time series & & 220050, 220179, \\
 & & & 224167, 225309, \\
        & & & 227243 \\
 Temperature & Time series & & 223761, 223762, \\
 & & & 224027 \\
 Weight & Time series & & 224639, 226512, \\
 & & & 226531 \\
 pH & Time series & & 220274, 220734, \\ &&& 223830, 228243 \\
    \midrule
	  \multicolumn{4}{c}{\bf Additional time series variables} \\
	  \midrule
	  Heart Rate & Time series & ICU Chartevents & 220045 \\
	  O2 saturation pulseoxymetry & Time series & ICU Chartevents & 220277 \\
	  Respiratory Rate & Time series & ICU Chartevents & 220210 \\
	  GCS - Eye Opening & Time series & ICU Chartevents & 220739 \\
	  GCS - Verbal Response & Time series & ICU Chartevents & 223900 \\
	  GCS - Motor Response & Time series & ICU Chartevents & 223901 \\
	  Alarms On & Time series & ICU Chartevents & 224641 \\
	  Parameters Checked & Time series & ICU Chartevents & 224168 \\
	  Heart Rate Alarm - Low & Time series & ICU Chartevents & 220047 \\
	  Heart rate Alarm - High & Time series & ICU Chartevents & 220046 \\
	  Non Invasive Blood Pressure mean & Time series & ICU Chartevents & 220181 \\
	  Non Invasive Blood Pressure systolic & Time series & ICU Chartevents & 220179 \\
	  Non Invasive Blood Pressure diastolic & Time series & ICU Chartevents & 220180 \\
	  O2 Saturation Pulseoxymetry Alarm - Low & Time series & ICU Chartevents & 223770 \\
	  O2 Saturation Pulseoxymetry Alarm - High & Time series & ICU Chartevents & 223769 \\
	  Resp Alarm - High & Time series & ICU Chartevents & 224161 \\
	  Resp Alarm - Low & Time series & ICU Chartevents & 224162 \\
	  Braden Sensory Perception & Time series & ICU Chartevents & 224054 \\
	  Braden Mobility & Time series & ICU Chartevents & 224057 \\
	  Braden Moisture & Time series & ICU Chartevents & 224055 \\
	  Braden Activity & Time series & ICU Chartevents & 224056 \\
	  Braden Nutrition & Time series & ICU Chartevents & 224058 \\
	  Braden Friction/Shear & Time series & ICU Chartevents & 224059 \\
	  SpO2 Desat Limit & Time series & ICU Chartevents & 226253 \\
	  Temperature Fahrenheit & Time series & ICU Chartevents & 223761 \\
	  IV/Saline lock & Time series & ICU Chartevents & 227344 \\
	  Gait/Transferring & Time series & ICU Chartevents & 227345 \\
	  Ambulatory aid & Time series & ICU Chartevents & 227343 \\
	  Mental status & Time series & ICU Chartevents & 227346 \\
	  Secondary diagnosis & Time series & ICU Chartevents & 227342 \\
	  History of falling (within 3 mnths) & Time series & ICU Chartevents & 227341 \\
	  Potassium (serum) & Time series & ICU Chartevents & 227442 \\
	  Sodium (serum) & Time series & ICU Chartevents & 220645 \\
	  Chloride (serum) & Time series & ICU Chartevents & 220602 \\
	  Creatinine (serum) & Time series & ICU Chartevents & 220615 \\
	  BUN & Time series & ICU Chartevents & 225624 \\
	  HCO3 (serum) & Time series & ICU Chartevents & 227443 \\
	  Anion gap & Time series & ICU Chartevents & 227073 \\
	  Hematocrit (serum) & Time series & ICU Chartevents & 220545 \\
	  Glucose (serum) & Time series & ICU Chartevents & 220621 \\
	  Hemoglobin & Time series & ICU Chartevents & 220228 \\
	  Platelet Count & Time series & ICU Chartevents & 227457 \\
	  WBC & Time series & ICU Chartevents & 220546 \\
	  Magnesium & Time series & ICU Chartevents & 220635 \\
	  Non-Invasive Blood Pressure Alarm - Low & Time series & ICU Chartevents & 223752 \\
	  Non-Invasive Blood Pressure Alarm - High & Time series & ICU Chartevents & 223751 \\
	  Phosphorous & Time series & ICU Chartevents & 225677 \\
	  Calcium non-ionized & Time series & ICU Chartevents & 225625 \\
	  Pain Level & Time series & ICU Chartevents & 223791 \\
	  Richmond-RAS Scale & Time series & ICU Chartevents & 228096 \\
	  Prothrombin time & Time series & ICU Chartevents & 227465 \\
	  INR & Time series & ICU Chartevents & 227467 \\
	  PTT & Time series & ICU Chartevents & 227466 \\
	  Capillary Refill R & Time series & ICU Chartevents & 223951 \\
	  Capillary Refill L & Time series & ICU Chartevents & 224308 \\
	  Admission Weight (lbs.) & Time series & ICU Chartevents & 226531 \\
	  Goal Richmond-RAS Scale & Time series & ICU Chartevents & 228299 \\
	  ST Segment Monitoring On & Time series & ICU Chartevents & 228305 \\
	  O2 Flow & Time series & ICU Chartevents & 223834 \\
	  Glucose finger stick (range 70-100) & Time series & ICU Chartevents & 225664 \\
	  Pain Level Response & Time series & ICU Chartevents & 224409 \\
	  Intravenous  / IV access prior to admission & Time series & ICU Chartevents & 225103 \\
	  20 Gauge Dressing Occlusive & Time series & ICU Chartevents & 227368 \\
	  Strength R Arm & Time series & ICU Chartevents & 228412 \\
	  Strength L Arm & Time series & ICU Chartevents & 228409 \\
	  Strength R Leg & Time series & ICU Chartevents & 228411 \\
	  Strength L Leg & Time series & ICU Chartevents & 228410 \\
	  20 Gauge placed in outside facility & Time series & ICU Chartevents & 226138 \\
	  Insulin pump & Time series & ICU Chartevents & 228236 \\
	  Self ADL & Time series & ICU Chartevents & 225092 \\
	  20 Gauge placed in the field & Time series & ICU Chartevents & 228100 \\
	  History of slips / falls & Time series & ICU Chartevents & 225094 \\
	  High risk (>51) interventions & Time series & ICU Chartevents & 227349 \\
	  Lactic Acid & Time series & ICU Chartevents & 225668 \\
	  Home TF & Time series & ICU Chartevents & 228648 \\
	  ETOH & Time series & ICU Chartevents & 225106 \\
	  Pressure Ulcer Present & Time series & ICU Chartevents & 228649 \\
	  Difficulty swallowing & Time series & ICU Chartevents & 225118 \\
	  18 Gauge Dressing Occlusive & Time series & ICU Chartevents & 227367 \\
	  18 Gauge placed in outside facility & Time series & ICU Chartevents & 226137 \\
	  Eye Care & Time series & ICU Chartevents & 225184 \\
	  Visual / hearing deficit & Time series & ICU Chartevents & 225087 \\
	  Currently experiencing pain & Time series & ICU Chartevents & 225113 \\
	  Dialysis patient & Time series & ICU Chartevents & 225126 \\
	  Daily Weight & Time series & ICU Chartevents & 224639 \\
	  Potassium & Time series & ICU Labevents & 50971 \\
	  Chloride & Time series & ICU Labevents & 50902 \\
	  Sodium & Time series & ICU Labevents & 50983 \\
	  Creatinine & Time series & ICU Labevents & 50912 \\
	  Urea Nitrogen & Time series & ICU Labevents & 51006 \\
	  Bicarbonate & Time series & ICU Labevents & 50882 \\
	  Anion Gap & Time series & ICU Labevents & 50868 \\
	  Glucose & Time series & ICU Labevents & 50931 \\
	  Hematocrit & Time series & ICU Labevents & 51221 \\
	  Platelet Count & Time series & ICU Labevents & 51265 \\
	  White Blood Cells & Time series & ICU Labevents & 51301 \\
	  Hemoglobin & Time series & ICU Labevents & 51222 \\
	  Red Blood Cells & Time series & ICU Labevents & 51279 \\
	  MCV & Time series & ICU Labevents & 51250 \\
	  MCH & Time series & ICU Labevents & 51248 \\
	  MCHC & Time series & ICU Labevents & 51249 \\
	  RDW & Time series & ICU Labevents & 51277 \\
	  Magnesium & Time series & ICU Labevents & 50960 \\
	  Phosphate & Time series & ICU Labevents & 50970 \\
	  Calcium, Total & Time series & ICU Labevents & 50893 \\
	  PT & Time series & ICU Labevents & 51274 \\
	  INR(PT) & Time series & ICU Labevents & 51237 \\
	  PTT & Time series & ICU Labevents & 51275 \\
	  pH & Time series & ICU Labevents & 50820 \\
	  Lactate & Time series & ICU Labevents & 50813 \\
	  Base Excess & Time series & ICU Labevents & 50802 \\
	  pO2 & Time series & ICU Labevents & 50821 \\
	  pCO2 & Time series & ICU Labevents & 50818 \\
	  Calculated Total CO2 & Time series & ICU Labevents & 50804 \\
	  \midrule
	  \multicolumn{4}{c}{\bf ICU static variables} \\
	  \midrule
	  Age & Numeric &  Admission Table & \\
	  Gender & Binary  &  Admission Table & \\
	  English Language & Binary  &  Admission Table & \\
	  Marital Status & Categorical  &  Admission Table & \\
	  Insurance & Categorical  &  Admission Table & \\
	  Admission Location & Categorical  &  Admission Table & \\
	  Admission Type & Categorical  &  Admission Table & \\
	  Race & Categorical  &  Admission Table & \\
	  First Care Unit & Categorical  &  ICU Admission Table & \\
	  Observation Window Length & Numeric & Derived & \\
   \midrule
	  \multicolumn{4}{c}{\bf ED vitals} \\
	  \midrule
   Temperature & Time series & Vital signs & \\
   Heart rate & Time series & Vital signs & \\
   Respiration rate & Time series & Vital signs & \\
   O2 Saturation & Time series & Vital signs & \\
   Systolic blood pressure & Time series & Vital signs & \\
   Diastolic blood pressure & Time series & Vital signs & \\
   \midrule
	  \multicolumn{4}{c}{\bf ED static variables} \\
	  \midrule
   	  Age & Numeric &  Patient table & \\
	  Gender & Binary &  Patient table & \\
   Temperature & Numeric & Triage & \\
   Heart rate & Numeric & Triage & \\
   Respiration rate & Numeric & Triage & \\
   O2 Saturation & Numeric & Triage & \\
   Systolic blood pressure & Numeric & Triage & \\
   Diastolic blood pressure & Numeric & Triage & \\
   Pain & Numeric & Triage & \\
    Acuity & Numeric & Triage & \\
	  \bottomrule
\caption{Time series and static variables used from MIMIC-IV dataset.}
\label{t:mimic-variables}
\end{longtable}}

\normalsize
For the ICU mortality task, our training set consists of a total of $19,699$ instances with a positive mortality rate of $12.95\%$, our validation set contains $4,257$ instances with a mortality rate of $13.55\%$, and our test set contains $4,245$ instances with a mortality rate of $12.39\%$.

Following \cite{nature_benchmark_2019}, the phenotyping task has 25 binary target variables, corresponding to whether the following conditions were billed during the stay:
\begin{itemize}[nosep,leftmargin=*] 
	\item Acute and unspecified renal failure
	\item Acute cerebrovascular disease
	\item Acute myocardial infarction
	\item Cardiac dysrhythmias
	\item Chronic kidney disease
	\item Chronic obstructive pulmonary disease and bronchiectasis
	\item Complications of surgical procedures or medical care
	\item Conduction disorders
	\item Congestive heart failure; nonhypertensive
	\item Coronary atherosclerosis and other heart disease
	\item Diabetes mellitus with complications
	\item Diabetes mellitus without complication
	\item Disorders of lipid metabolism
	\item Essential hypertension
	\item Fluid and electrolyte disorders
	\item Gastrointestinal hemorrhage
	\item Hypertension with complications and secondary hypertension
	\item Other liver diseases
	\item Other lower respiratory disease
	\item Other upper respiratory disease
    	\item Pleurisy; pneumothorax; pulmonary collapse
	\item Pneumonia (except that caused by tuberculosis or sexually transmitted disease)
	\item Respiratory failure; insufficiency; arrest (adult)
	\item Septicemia (except in labor)
	\item Shock
\end{itemize}
This task has a total of $54,024$ training instances, $11,401$ validation instances and $11,509$ testing instances.

The ED transfer to ICU task has a total of $91,479$ training instances, $20,171$ validation instances, and $19,660$ testing instances.

\end{document}